\documentclass[10pt]{iopart}

\usepackage[margin=2.2cm]{geometry}

\usepackage{amsfonts}
\usepackage{amssymb}
\expandafter\let\csname equation*\endcsname\relax
\expandafter\let\csname endequation*\endcsname\relax
\usepackage{amsmath}
\usepackage{multirow}
\usepackage[pdftex]{graphicx}

\newtheorem{mydef}{Definition}

\newcommand{\argmin}{\arg\!\min}

\usepackage{amsopn}
\DeclareMathOperator{\rank}{rank}

\usepackage[noend]{algpseudocode}
\usepackage{algorithmicx}

\begin{document}

\title[]{Robust Cardiac Motion Estimation using  Ultrafast Ultrasound Data: A Low-Rank-Topology-Preserving Approach}

\author{Angelica I. Aviles$^1$, Thomas Widlak$^2$, Alicia Casals$^{1,5}$, Maartje M. Nillesen$^3$ and Habib Ammari$^4$}

\address{$^1$ The Research Center of Biomedical Engineering (CREB), Universitat Polit\`{e}cnica de Cataluya, Spain.\\
 $^2$ The Department of Applied Mathematics and Applications, \'{E}cole Normale Sup\'{e}rieure, Paris, France. \\
 $^3$ The Medical UltraSound Imaging Center, Radboud University Nijmegen Medical Centre, Nijmegen, The Netherlands.\\
 $^4$ The Department of  Mathematics, Eidgen\"ossische Technische Hochschule Z\"urich - ETH Zurich, Switzerland \\
 $^5$ Institute for Bioengineering of Catalonia, Barcelona, Spain}
\ead{angelica.ivone.aviles@upc.edu}
\vspace{10pt}

\begin{abstract}
Cardiac motion estimation is an important diagnostic tool to detect heart diseases and it has been explored with modalities such as MRI and conventional ultrasound (US) sequences. US cardiac motion estimation still presents challenges because of the complex motion patterns and the presence of noise. In this work, we propose a novel approach to estimate the cardiac motion using ultrafast ultrasound data. -- Our solution is based on a variational formulation characterized by the $L^2$-regularized class. The displacement is represented by a lattice of b-splines and we ensure robustness, in the sense of eliminating outliers, by applying a maximum likelihood type estimator. While this is an important part of our solution, the main highlight of this work is to combine a low-rank data representation with topology preservation. Low-rank data representation (achieved by finding the $k-$dominant singular values of a Casorati Matrix arranged from the data sequence) speeds up the global solution and achieves noise reduction. On the other hand, topology preservation (achieved by monitoring the Jacobian determinant) allows to radically rule out distortions while carefully controlling the size of allowed expansions and contractions. Our variational approach is carried out on a realistic dataset as well as on a simulated one. We demonstrate how our proposed variational solution deals with complex deformations through careful numerical experiments. The low-rank constraint speeds up the convergence of the optimization problem while topology-preservation ensures a more accurate displacement. Beyond cardiac motion estimation, our approach is promising for the analysis of other organs that experience motion.
\end{abstract}

%
%
%
%
%
\section{Introduction}\label{sec:introduction}

Acording to the World Health Organization, cardiovascular diseases are the leading cause of death in the world. A key factor for early detection and prevention of these diseases is to analyze the cardiac motion to diagnose, for example, valve conditions and motion abnormality.  Moreover, the cardiac mechanics can be studied and analyzed through the heart deformation \cite{Duchateau::2013}. In most of the medical laboratories, diagnosis is based on the visual inspection of the heart's motion \cite{Hoit::11}. However, the results are conditioned to the experience of an expert, and in consequence, they are highly variable and subjective. Thereby, the need of having objective and understandable measures emerge, and up-to-date cardiac motion estimation is a central topic in cardiac imaging \cite{lopez::15}.

Estimation of cardiac motion is a very active pro\-blem to be tackled. In the search of achieving a good motion estimation, different authors have used diverse biomedical imaging modalities including: magnetic resonance imaging (MRI), computed tomography (CT), single photon emi\-ssion computed tomography (SPECT), and positron emi\-ssion tomography (PET)  (e.g. \cite{Prummer::09,Wang::15,Royuela::15}). However, the lack of resolution in modalities such as PET and SPECT ($\approx 4-7$~mm \cite{Nappi::15}), together with the exposure to radiation (i.e. CT, PET/SPECT), and the magnetic interference and cost of MRI make their use complicated.

An alternative modality to estimate cardiac motion is ultrasound imaging (US). US is very popular due to its low cost, high accessibility, real-time interaction, non-ionization, portability, and rapid assessment  \cite{Spencer::2015}. It has become routinely used  in multiple clinical scenarios including diagnostic and heart disease prevention. US a\-llows capturing, for example, the heart's size and shape, strain rate, ventricular deformation, and abnormal motions. Moreover, its feasibility for tissue tracking and estimated motion analysis has been demonstrated \cite{DeLuca::15,Shung::2015}.

Despite these benefits, ultrasound has disadvantages related to the presence of noise and occasional artifacts and as well a limited acquisition speed. The poor temporal resolution of conventional US hinders retrieving different mechanical events of the heart \cite{Cikes::14}, \cite{Tanter::14}. Nonetheless, recent advances in ultrafast ultrasound (UUS) imaging have overcome previous issues, particularly temporal resolution, as they have a much higher frame rate (greater than 1000fps) \cite{Tanter::15}, which is advantageous for cardiac motion estimation. An UUS system is capable of computing many lines in parallel, generating in this way a full image from one single transmitting event. Different applications of UUS emerge, such as tissue and blood motion estimation and imaging of micro bubbles or neurovascular coupling. Moreover, it facilitates advancements in disease prevention, diagnosis, and therapeutic monitoring \cite{Tanter::14}. In this work, we study the application of UUS to estimate the cardiac motion.

In the literature, different works for US cardiac motion estimation have been reported. Most of them use conventional ultrasound (e.g. \cite{ledesma::05,Zhang::12,Heyde::2016,Zhang::14}), while few works have been reported using ultrafast ultrasound imaging (e.g. \cite{Nillesen::15,Salles::15}).  Despite the use of conventional or ultrafast ultrasound, the approaches proposed to retrieve the cardiac motion can be classified into those using the characteristics of the radio frequency (RF) signal, and those based on computer vision techniques applied to an image sequence.

Motion estimation techniques in the first category use the natural acoustic reflections of the radio frequency (RF) signal. When this option is used, the cardiac motion can be  computed by either applying speckle tracking techniques, which use the amplitude of the signal, or radio-frequency-based correlation techniques, which use the phase information  (e.g. \cite{Salles::15,Dydenko::03,Lopata::11}). This option is a promising approach to estimate the cardiac motion. However, the result greatly depends on the characteristics of the RF.


In the second category of motion estimation techniques, we can find solutions based on Optical Flow (OF), in which the relative motion of the heart is computed from the velocities of patterns' brightness (examples can be found in \cite{Tenbrinck::13,Vyas::14,Gao::15}). Although different authors have proved the feasibility of OF for motion estimation, its main drawback is that it only works with relatively non-complex deformations.  Another common solution, usually adapted for more complex motions, is non-rigid registration, in which displacements of the tissue can be tracked by computing the spatial correspondences between frames \cite{ledesma::05,Metz::11, Oktay::15, Morin::15}.

Another option for estimating the dynamics of the heart is by tracking the heart's borders using deformable models
(e.g. \cite{Ahn::12,Huang::14,Milletari::15}). However, cardiac motion estimation can be inaccurate when the displacements are parallel to the edge or when there is a lack of well-defined borders, which is a common problem when using US \cite{ledesma::05}.
\medskip

\begin{figure*}[!t]
\centering
\includegraphics[width=17cm]{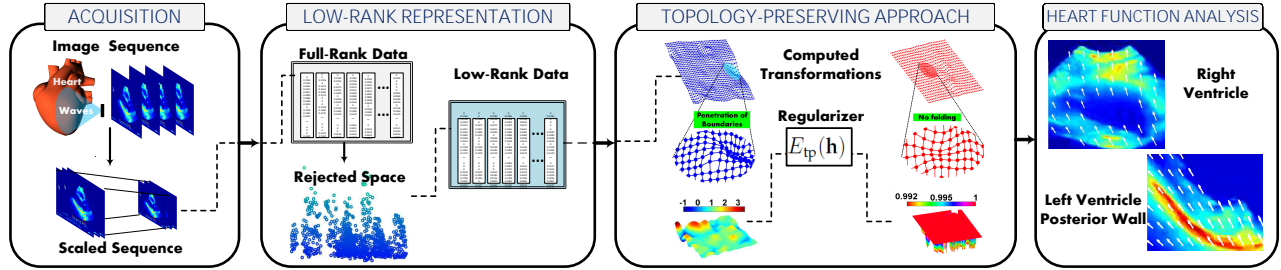}
\caption{Overview of our proposed approach. (From left to right) an ultrafast ultrasound cardiac sequence is first acquired, this data is then represented in low-rank in order to speed up the solution and reduce noise. Later, cardiac motion is computed enforcing topology preservation which allows keeping the anatomical structure of the heart. Finally, analysis of the results is offered.}
\label{fig:ValueChain}
\end{figure*}

In this work, we propose a new approach to estimate the cardiac motion using ultrafast ultrasound modality (see Figure.~\ref{fig:ValueChain}). Based on a variational formulation for non-rigid registration in $L^2$, we include a maximum likelihood type estimator to increase  the robustness of the solution in the sense of being able to deal with outliers. While this is an important part of the solution, the main contribution is: combining low-rank data representation with a topology-preserving approach. Particularly:

\begin{itemize}
\item We promote low-rank data representation. As a stand-alone tool, it already offers  several advantages, such as speeding up the global solution by reducing the computational time and decreasing the noise in the image sequence.
\item Another key point is topology preservation. A penalization term for the Jacobian determinant is used in order to guarantee a diffeomorphic transformation. We use a regularizer to rule out distortions while controlling the magnitude of expansion and compression.
\item The combination of the two previous tools turns out to be powerful as it allows computing an accurate displacement field that is mathematically well-motivated and computationally efficient.
\end{itemize}

The remainder of this paper is organized as follows. Section~\ref{sec:Related} presents previous literature related to the topology-preserving problem. Our solution is tackled in Section~\ref{sec:Methodology}. Particularly, in Subsection \ref{subsec:LowRank} we present the low-rank data representation strategy,  while in Subsection~\ref{subsec:Variational}
we describe our variational framework to recover the deformation over time. Moreover, in Subsection~\ref{subsec:Top} we describe our penalization term for achieving topology preservation. In Section~\ref{sec:Experiments}, we validate our proposal presenting experiments on both, a realistic and a simulated dataset. Section 5 provides a final conclusion and directions for future works.

\section{Related Work}
\label{sec:Related}

During cardiac motion estimation, unpreserved topologies result in violations of region convexity and are reflected in penetration of boundaries and overlapped or distorted mesh elements. Thereby, topology preservation is important in order to ensure connectivity between structures, maintain the relations between neighboring elements, and avoid distortion of existing structures.  When the deformations are small, topology is preserved by the smoothness offered by the regularization term, but this is not the case when large deformations appear, as it happens when the complex dynamics of the heart is to be retrieved.

From the conservation principles of continuum mechanics, it is clear that the elastic displacement field in cardiac motion can be modeled as isomorphism resp. diffeomorphism. Necessary and sufficient conditions  are that its deformation gradient tensor exists and is nonsingular at every point in the deformable body \cite{Slaughter::02}. In terms of the Jacobian $J$, this means that $J$ exists and that the $\det(|J|)\neq 0$ at every point in the deformable body. For a positive volume, it is required that $|J|>0$ throughout the deformable body \cite{Slaughter::02}.

A well-known approach to achieve topology preservation is controlling the Jacobian determinant. Dacorongna in  \cite{Dacorogna::90} presented a detailed discussion related to the Jacobian determinant equation in which he demonstrated its ability to achieve topology preservation.
Jacobian determinant has been used in problems involving deformable structures in order to achieve more realistic transformations (e.g. \cite{Christensen::96,Ashburner::99,Ozere::15}).  However, in this section we  cover only those that have relation with the modeling of deformable objects. In a multidimensional elastic registration framework, authors in \cite{Kybic::00} explored a barrier function for penalizing locally non-invertible functions.

The problem of achieving  topology preservation has been reported in different works addressing MRI and US data.
The authors in \cite{Christensen::96} ensured topology preservation by defining a threshold of 0.5 for the Jacobian determinant. Then, for values lower than this threshold, they generate a new template, equal to the previous deformed template, to continue with the registration process.
Similarly, Ashburner et al. in \cite{Ashburner::99}, \cite{Ashburner::00} penalized singular values of the Jacobian having lognormal distribution.  To enforce the Jacobian positivity, Musse and colleagues \cite{Musse::01} proposed a {2D} parametric approach based on the constraint of the continuous hierarchical modeling of the deformation field.

Later on, Noblet et al.  \cite{Noblet::05} reported an extension of Musse's work in the three-dimensional space. They presented a hierarchical  deformation field model in which the Jacobian determinant was conditioned when it had negative values. The main difference between the two works is that the nontrivial optimization problem in \cite{Noblet::05} was obtained in the 3D space.
Topology preservation was treated as a hard constraint in \cite{Davatzikos::04},  where they restricted the Jacobian determinant within a set of intervals in a grid region. If these conditions were not met, then they enforced topology preservation in terms of gradients.  Explicit control of the deformation in terms of the determinant of the Jacobian was reported in \cite{Haber::2007}. In comparison with similar works, here, the authors proposed the use of point-wise inequality constraints (i.e. they achieved topology preservation by controlling voxel by voxel instead of using integral measures).

Another approach was presented in \cite{Yanovsky::07}, in which topology was preserved by quantifying the magnitude of deformations and examining the statistical distributions of Jacobian maps in the logarithmic space. A two-step solution was proposed in \cite{Guyader::12}. Authors first corrected the gradient vectors of the deformation and then they reconstructed the deformation based on a minimization problem on a convex subset of the underlying Hilbert space. As a result, they achieved a well-defined Jacobian in the image domain. An extension of that work was presented in \cite{Ozere::15}, in which authors proposed a solution based on independent problems of small dimension that allow parallel computation.

Zhang et al. \cite{Zhang::12,Zhang::14} developed a temporally diffeomorphic motion estimation approach for conventional cardiac ultrasound sequences. In that work, the authors addressed the topology-preservation problem by using a smooth velocity field with a differential operator in a Sobolev space. The resulting transformation defines a group of diffeomorphisms.
In \cite{Lui::15}, authors used the Beltrami coefficient (BC) to represent an orientation-preserving diffeomorphism. To deal with the computational cost of the BC method, they presented a splitting algorithm, one part solved the BC whereas the other involved the quasi-conformal map. However, the BC was reduced to constrain the Jacobian. Moreover, a biophysically constrained framework for large deformation diffeomorphic image registration was proposed in \cite{Mang::15}. They achieved topology preservation by controlling the Jacobian determinant and the amount of shear in the deformation map using a nonlinear Stroke regularization scheme. 

\section{Cardiac Motion Recovery}
\label{sec:Methodology}
In this section, we present our solution to estimate the cardiac motion. The section contains three parts: First we describe the low-rank representation of the data. Then we give the variational framework to recover the deformation field. Finally, we present our regularizer for topology preservation to obtain a realistic diffeomorphic solution.

\subsection{Low-Rank Data Representation}
\label{subsec:LowRank}
As a recent promising tool, low-rank data representation has been promoted in a variety of areas, including but not limited to: computer vision, machine learning  for fitting problems, and computing the low-rank approximation of a matrix \cite{Candes::13, Newson::15}. Low-rank representation has been  useful for processing big data because in many datasets, the relevant information lies in a low-dimensional space \cite{Haeffele::14}. In particular, low-rank representations have been used in motion recovery in optical flow, e.g. in \cite{Agapito::13}, to recover the motion in videos of facial movements.

Our motivation for using low-rank data is threefold: First, we aim to increase the computational speed of the solution convergence. Second, it constitutes a means of denoising
the ultrasound data and avoiding artifacts in the recovered deformation field. Third, we aim to investigate the synergy of the low-rank representation with the preservation of the topology.

For a precise description of our  low-rank data representation, consider
an ultrafast ultrasound image sequence, $F=\lbrace f_{s} \rbrace_{s=0}^{S-1}$, with $S$ frames of size $M*N$. Then, a new structure of the data is given in the form of a single matrix, called Casorati matrix $\textbf{C}$, which columns are the $S$ frames in a vectorized way:
\begin{equation}
\label{eq:Casorati}
\textbf{C}=\textbf{C}(F) = \left[
\begin{array}{c l}
     f_{0}(1,1)   & \cdots f_{S-1}(1,1)\\
      \vdots	&   \text{  \hspace{1cm}     }   \vdots \\
      f_{0}(M,N) &  \cdots f_{S-1}(M,N)\\
\end{array}\right]
\end{equation}
where $f_{s}(m,n)$ is the scalar value of the sequence in frame $s$, at a given pixel location $(m,n)$.

\begin{table}[!t]
\centering
\caption{Decomposition of the Casorati matrix $\mathbf C$ and the rank 100-approximation $\mathbf C_{100}$}
\label{tab:SVD}
\begin{tabular}{|l|l|l|l|}
\hline
      & \multicolumn{1}{c|}{\textbf{U}} & \multicolumn{1}{c|}{\textbf{S}} & \multicolumn{1}{c|}{\textbf{V}} \\ \cline{1-4}
$\mathbf C$  & 13824x13824  & 13824x1814  & 1814x1814  \\ \cline{1-4}
$\mathbf C_{100}$   & 13824x100    & 100x100     & 1814x100   \\ \hline
\end{tabular}
\end{table}

\begin{figure}[!t]
\centering
\includegraphics[width=9cm]{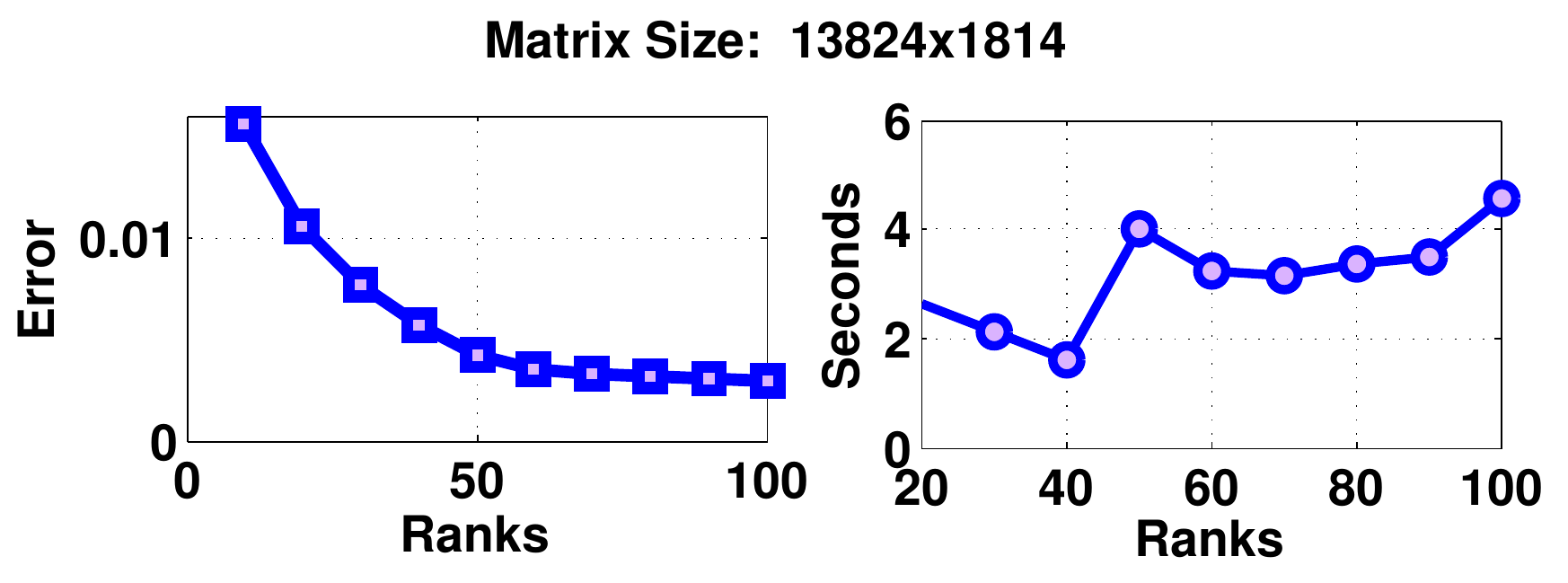}
\caption{Left: Singular values $\sigma_{k+1}=\| \mathbf C-\mathbf C_k \|^2$. Right: CPU time to compute the rank-$k$ approximation $\mathbf C_k$. }
\label{fig:RankCasorati}
\end{figure}

Let us now turn to the theoretical prerequisites to produce a low-rank representation of the Casorati representation $\textbf C$, exploiting the high correlation in the columns of the matrix.

It is well-known (see, for instance, \cite{Strang::05}) that any matrix $\textbf{A}\in \mathbb{R}^{M*N}$ admits a decomposition in the way:

\begin{equation}
\label{eq:SVD}
\textbf{A}=\sum_{y=1}^{r}\sigma_{y}\textbf u_{y}\textbf v_{y}^{T}=\textbf{U} \textbf{S} \textbf{V}^{T}
\end{equation}
where $\textbf{U}=(\textbf u_1,\ldots,\textbf u_M)\in \mathbb{R}^{M*M}$ and $\mathbf{V}=(\textbf v_1,\ldots,\textbf v_N) \in R^{N*N}$ are orthogonal matrices;  the matrix $\textbf{S}$  is diagonal $\textbf{S}=(\sigma_{1},...,\sigma_{r},0,...,0)$ with $r\leqslant \min(M,N)$ and the scalar quantities $\sigma_{i}$ are called the singular values of the matrix $\textbf{A}$. Then, the triplet $\textbf{U} \textbf{S} \textbf{V}$ is called a singular value decomposition (SVD).

A major issue in medical applications is the large amount of data to be processed, which is the case
when the cardiac motion is estimated. An option to deal with this issue is to compute the set of dominant singular values instead of computing the singular value
decomposition of a large and dense matrix. This allows keeping the most relevant information in a subspace which is smaller than the original one. Thus, the problem of building a  low-rank representation can be  given by finding the $k-$dominant singular values of $\mathbf{A}$. Mathematically,  finding  the rank-$k$ approximation of the matrix  $\mathbf{A}$ can be described as:
\begin{equation}
\label{eq:rankK}
\mathbf{A}_{k}:=\sum_{y=1}^{k}\sigma_{y}\textbf u_{y}\textbf v_{y}^{T}= \argmin_{\rank(\mathbf{\widehat{A}} )\leq k} \| \textbf{A}-\mathbf{\widehat{A}} \|_{F}^{2}
\end{equation}
where $\| \mathbf L \|_{F}^{2}$ is the squared Frobenius norm of a matrix. It is used in low-rank based problems since it is invariant to rotations and to the rank. The importance of the rank-$k$-approximation in \eqref{eq:rankK} is given by the Eckhart-Young theorem (see, for instance,  \cite{Golub::96}).
Take a matrix $\mathbf A$ with a SVD as in \eqref{eq:SVD}, and let $k<r:=\rank(\mathbf A)$. Let $\mathbf A_{k}$ be the rank-$k$ approximation in equation \eqref{eq:rankK}. Then, we have:
\begin{equation}
\label{eq:EckhardYoung}
	\| \mathbf A-\mathbf A_{k} \|^{2} = \min_{\rank(\mathbf{B} )= k}\| \mathbf A-\mathbf B \|^{2}=\sigma_{k+1}
\end{equation}
\medskip

For our data described in Section \ref{sec:Experiments}, we compute the Casorati matrix $\mathbf C(F)$ and different rank-$k$-approximations $\mathbf C_k$ in equation \eqref{eq:rankK}. The resulting matrix decomposition is displayed in Table \ref{tab:SVD}, in which the original matrix (of size 13824x1814) was reduced to rank 100. Both the error and the computational time are plotted against the rank $k$ in Figure \ref{fig:RankCasorati}. This figure shows that only an average of $2.08984$ seconds was needed to obtain $\mathbf{C}_{k}$, and that the approximation error for $\mathbf C_{100}$ is $1.44e^{-5}$.


\begin{figure}[!t]
\centering
\includegraphics[width=11.5cm]{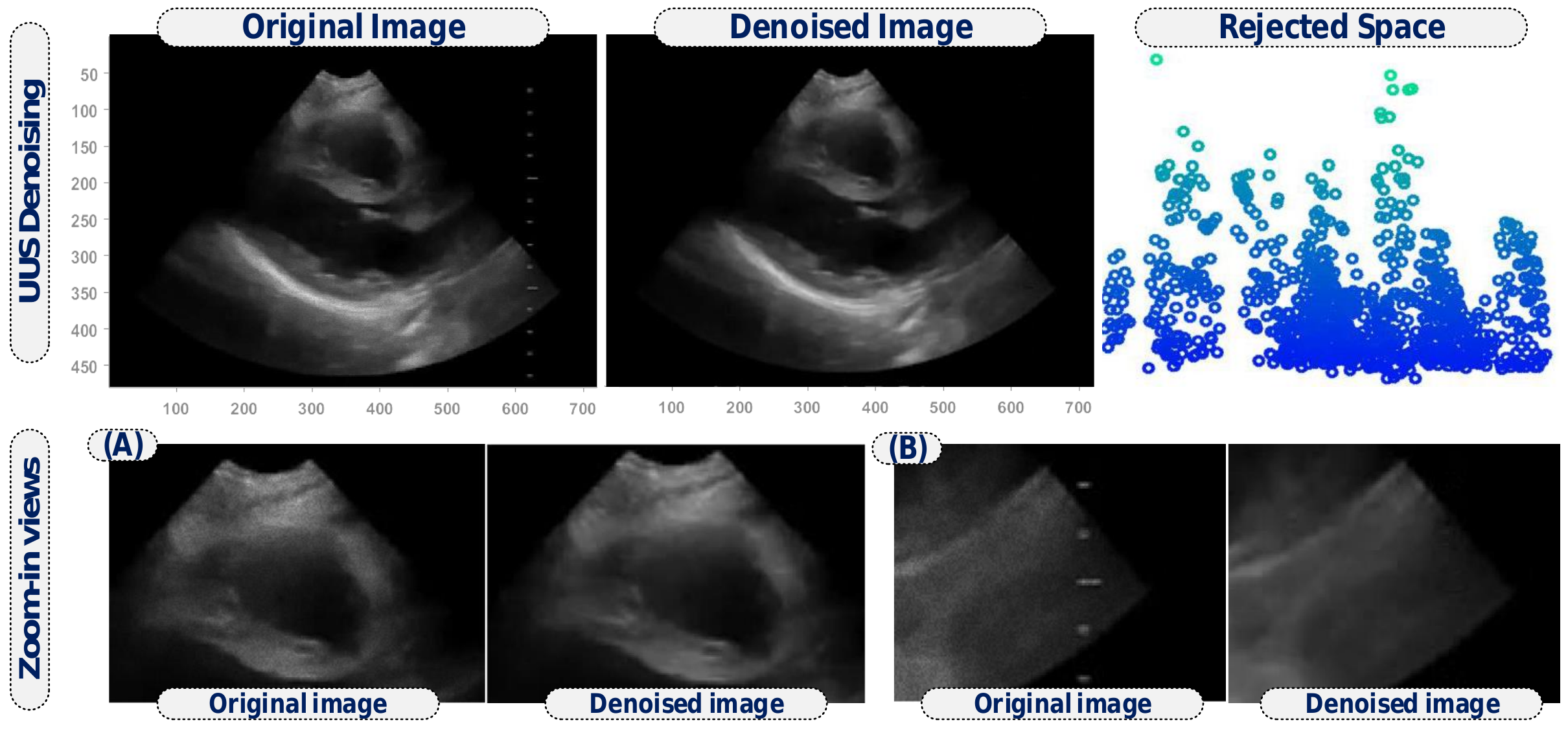}
\caption{Top row shows an original and denoised frame after applying low-rank process along with the rejected space. Bottom row, A and B, show zoom in views of the same frames in which we can see that both noise and some artifacts were removed. }
\label{fig:Noise}
\end{figure}

As we stated before, another main motivation to promote low-rank representation $\mathbf C_k$ instead of using the full Casorati matrix $\mathbf C$ is to reduce noise.
This is accomplished by eliminating the subspace where the noise relies, which results in retrieving a subspace with only relevant information.  Noise, which normally relies on another
subspace due to its characteristics, is rejected from the solution (see Figure~\ref{fig:Noise}). This eliminates artifacts in the subsequent deformation computation.

In the last step, we use the invertibility in equation~\eqref{eq:Casorati} to invert the low-rank representation $\mathbf C_k(F)$ back to the denoised video sequence $F_k=((f_s)_k)_{s=0}^{S-1}$. We will work with that sequence in the following section to extract the mechanical deformation field of the heart.

Let us now expose the variational framework to recover the deformation.

\subsection{Deformation Recovery}
\label{subsec:Variational}
Let $F = \{ f_s\}_{s=0}^{S-1}$ be the image sequence over $S$ frames, where each image $f_s$ is a function over the bounded domain~$\Omega$.

We will find a deformation vector field $\mathbf h$ defined on the domain $\Omega$ as a minimizer of an energy functional:
\begin{equation}
\label{eq:Energy}
E(\mathbf h)=\sum_{s=0}^{S-2}E_{\text{dsc}}(f_s , f_{s+1};\mathbf h)+E_{\text{reg}}(\mathbf h) + E_{\text{tp}}(\mathbf h)
\end{equation}
where the three terms used have the following purposes: $E_\text{dsc}$ discrepancy measure, $E_\text{reg}$ regularization term and $E_\text{tp}$ topology preservation.

We will now go through our variational framework and explain each of the terms in the energy functional. We begin with the representation of the deformation field $\mathbf h$, then go on to the discrepancy term $E_\text{dsc}$ and the regularization term $E_\text{reg}$. In Subsection \ref{subsec:Top}, we describe the topology preservation term $E_\text{tp}$ necessary to achieve realistic deformations.


How to represent the deformation field $\mathbf h$? The deformation model is an essential factor that defines how fast and accurate the approach is. In order to find a compromise between computational cost and accuracy, we will handle the changes over time using a lattice as in the following definition:
\begin{mydef}
A \emph{m-dimensional lattice} is  the $\mathbb{Z}-$linear span  of a set of $k$ linearly independent vectors in $\mathbb{R}^{m}$.
\end{mydef}
We will then use a lattice in which its points are characterized by the tensor product of the b-splines \cite{Unser:99}. These are widely used in medical applications. The advantage of this lattice deformation model is that it demands low running time, allows multiresolution, have optimal mathematical properties, and keeps affine invariance. Using b-splines has the additional advantage of being able to handle complex deformations.

Consider a given position $\textbf w=(w_1,\ldots,w_d)$ in~$\mathbb{R}^d$. Let $\{\xi_i(\cdot)\}$ be a basis of spline functions and let $\mathbf{P}_{j_1,\ldots j_d}$ be control points.  Then, we express the deformation vector $\mathbf h$ at point $\mathbf w$ through the model:

\begin{equation}
\begin{aligned}
\mathbf h(\mathbf w)=\sum_{j_{1}=0}^{n}...\sum_{j_{d}=0}^{n}\overbrace{\mathbf{P}_{j_{1},...,j_{d}}}^{control\text{ } points}\underbrace{\prod_{k=1}^{d}\xi_{j_k}( w_{k})}_{tensor\text{ }product}
\label{eq:LatticeDeformation}
\end{aligned}
\end{equation}
In this work, we use cubic basis splines:
\begin{equation}
\label{eq:Splines}
\xi_{0}(x)=(1-x)^{3}/6, \text{   }\text{   } \xi_{1}(x)=(4-6x^{2}+3x^{3})/6, \text{   }\text{   } \xi_{2}(x)=(1+3x+3x^{2}-3x^{3})/6, \text{   }\text{   } \xi_{3}(x)=x^{3}/6
\end{equation}
The deformation model for $\mathbf h$ in \eqref{eq:LatticeDeformation} is in $\mathbb{R}^d$ and it shows that within our framework, we actually reconstruct the control points $\mathbf P_{j_1,\ldots,j_d}$ in order to get the deformation $\mathbf h$. -- In our application, we will exploit this deformation model for dimension $d=2$.

\medskip

We now turn to the discrepancy term $E_{\text{dsc}}$. Since the images are acquired by the same sensor, it is not expected that they have a big intensity variation between them. Therefore, an iconic method is a perfect match for this application. One possible option is to use the Sum of Squared Differences (SSD) method:
\begin{equation}
\label{eq:SSD}
	\int_{\Omega}f_0(\textbf w+\textbf h(\textbf w) - f_1)^2 d\textbf w
\end{equation}
SSD offers a low computational cost but it has the disadvantage of not dealing well with outliers.

For the actual expression for $E_{\text{dsc}}$ in \eqref{eq:Energy}, we use:
\begin{equation}
	E_{\text{dsc}}(f_0,f_1;\textbf h)=\int_{\Omega}\rho\left(f_0(\textbf w+\textbf h(\textbf w)) - f_1\right) d\textbf w
\end{equation}
Here, the function $\rho$ is a maximum likelihood type estimator motivated by robust statistics:

\begin{mydef}An \emph{M-estimator} is a symmetric and positive definite function $\rho$ with a unique minimum at zero.
\end{mydef}

The M-estimator substitutes the minimization of  $\sum_{i}\mathbf{r}_{i}^2$, where $\mathbf{r}$  is the residual error, with $\sum_{i}\rho(\mathbf{r}_{i})$, in order to deal with the effect of outliers. This increases the robustness and accuracy of the result.

In this work, we use the Turkey estimator for $\rho$:
\begin{equation}
\label{eq:Tukey}
\rho(x)=\left\{
\begin{array}{c l}
    \frac{c^2}{6} \Big[1-\Big(1-(\frac{x}{c})^2\Big)^3\Big] & \text{if }| x|\leq c\\
    \frac{c^2}{6} &\text{if } |x|> c
\end{array}\right.
\end{equation}
where $c$ is a tunning parameter.
The discrepancy term with the Turkey estimator in \eqref{eq:Tukey} is known to offer a hard rejection of outliers \cite{Stewart::99}.

\medskip

For the regularization term $E_{\text{reg}}$, we use the Tikhonov method to impose stability to the energy functional. Let $\gamma\in \mathbb{R}^{+}$ be the regularization parameter. Then we use for $\textbf  h=(h_1,\ldots,h_d)$ the term:
\begin{equation}
E_{\text{reg}}(\textbf h)= \gamma \sum_{l=1}^{d} \int_{\Omega} \|\nabla h_{l}(\textbf w)  \|^2 d\mathbf w
\end{equation}
as the regularizer in our variational framework \eqref{eq:Energy}.

\subsection{Topology Preservation}\label{subsec:Top}
\begin{figure}[!t]
\centering
\includegraphics[width=9cm]{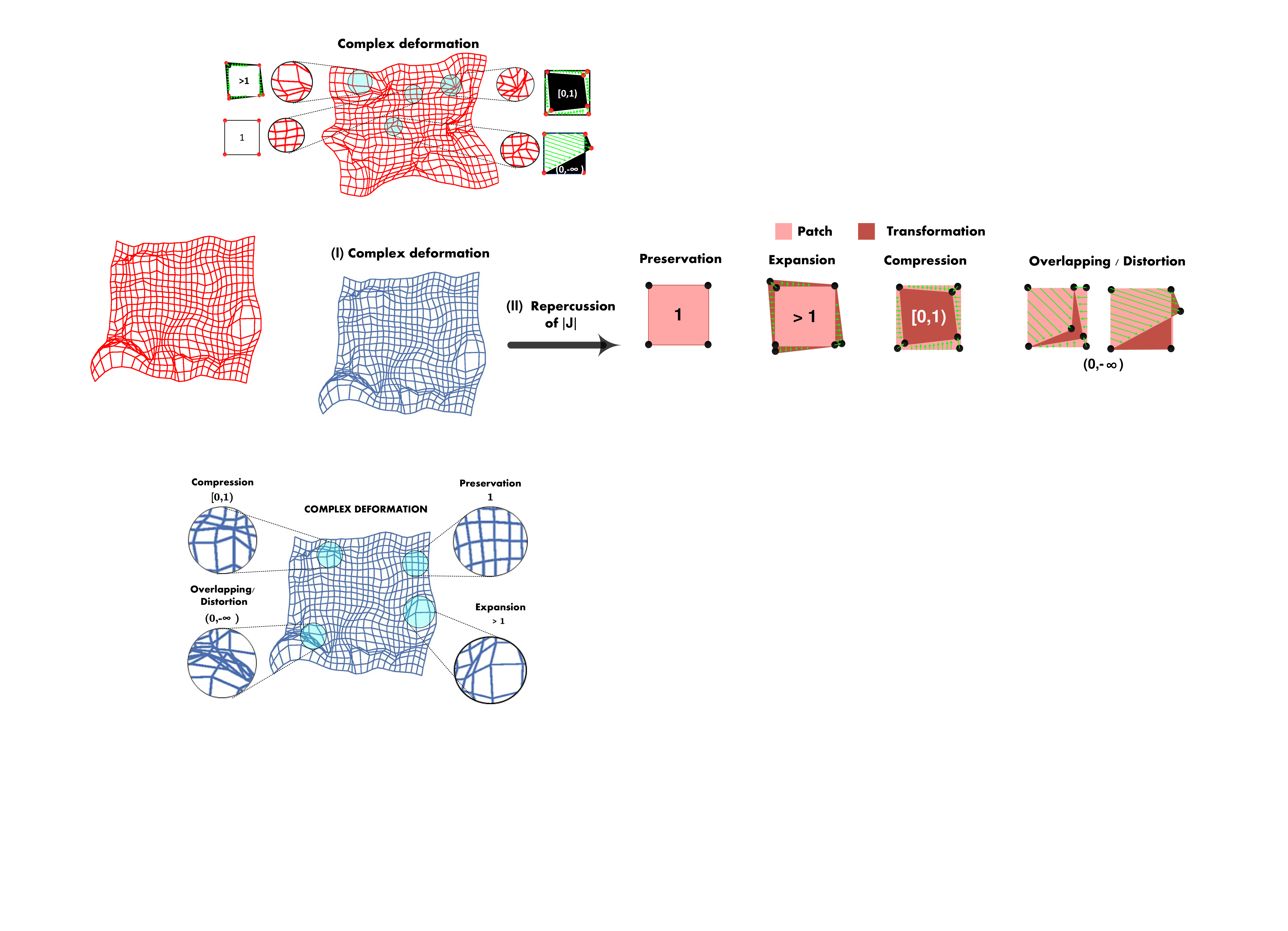}
\caption{When topology-preserving is not enforced, unrealistic transformations result. A way to ensure topology preservation is by checking the Jacobian determinant $|J|$. When it is equal to 1 then the volume is preserved. Small positive or large positive numbers of $|J|$ result in contractions or expansions. But having $|J|\in(-\infty,0)$ can result in distortions, overlapping, and creation of new structures.}
\label{fig:TopologyBehavior}
\end{figure}

A common drawback of most of the solutions working with complex deformations is that
topology-preserving is not guaranteed, which leads to unrealistic deformations becoming a source of error. An example of such deformations is shown in Figure \ref{fig:TopologyBehavior} in which topology is not preserved. In medical applications, this issue is of huge importance in order to maintain the anatomical structures.  In order to preserve the topology, the deformation must be a diffeomorphism. One way to achieve topology preservation is using the Jacobian determinant. This option makes sense since it allows measuring the changes in the area/volume produced by the deformation at each patch.

\begin{table}[!t]
\begin{flushright}
\resizebox{14cm}{!} {
\begin{tabular}{|c|p{11cm}|}
\hline
Condition & Local type of deformation \\
\hline
$| J_{\mathbf h}(\mathbf w)|\leq 0 $&  topology is destroyed, overlapping, distortion and penetration of boundaries may occur \\ \hline
	$| J_{\mathbf h}(\mathbf w)|>0 $& diffeomorphism, topology preservation \\ \hline
	$0<| J_{\mathbf h}(\mathbf w)|<1 $&  contraction\\  \hline
	$| J_{\mathbf h}(\mathbf w)| = 1$& volume preservation \\	\hline	
	$1<| J_{\mathbf h}(\mathbf w)|$ & expansion \\ \hline	
\end{tabular}}
\end{flushright}
\caption{Jacobian determinant conditions for topology preservation}
\label{tab:conditionsJ}
\end{table}

Let $\vert J_{\mathbf h}(\mathbf w)\vert$ be the Jacobian determinant of the deformation $\mathbf h=(h_x,h_y)$ in $\mathbb{R}^2$. The Jacobian determinant is described as:

\begin{equation}
\label{eq:Determinant}
|J_{\textbf h}(\textbf w)|=\det \left(
\begin{array}{c l}
     \frac{\partial h_{x}(\textbf w)}{\partial x}  \text{    } \frac{\partial h_{x}(\textbf w)}{\partial y} \\
     \frac{\partial h_{y}(\textbf w)}{\partial x}  \text{    } \frac{\partial h_{y}(\textbf w)}{\partial y}  \\
\end{array}\right)
\end{equation}

Table 2 shows the characteristics of the deformation~$\mathbf h$ encoded in the Jacobian and an illustration of these behaviors can be seen in Figure~\ref{fig:TopologyBehavior}.

From our deformation model in \eqref{eq:LatticeDeformation}, the partial derivatives of $\frac{\partial h_{x}(\textbf w)}{\partial x},...,\frac{\partial h_{y}(\textbf w)}{\partial y} $ can be easily evaluated, as they come from the tensor product of independent functions. Using Equation \eqref{eq:Splines}, we have for the derivatives of the cubic basis splines:
\begin{equation}
\begin{aligned}
\xi_{0}'(x)=(1-x)^{2}/2, \text{ } \text{   }\text{   } \xi_{1}'(x)=(3x^{2}-4x)/2, \text{   }\text{   } \xi_{2}'(x)=(-3x^{2}+2x+1)/2, \text{   }\text{   } \xi_{3}'(x)=x^{2}/2
\label{eq:2}
\end{aligned}
\end{equation}
Then, the determinant in \eqref{eq:Determinant} can be straightforwardly evaluated as a function of the lattice points $\mathbf P_{j_1,\ldots,j_d}$ characterizing the deformation $\mathbf h$.

\medskip

Once we have stated how to compute the Jacobian determinant, we turn to formulate our penalization term $E_{tp}$ for the energy functional \eqref{eq:Energy}. We use a weak constraint, but do not penalize values lying near 1:
\begin{equation}
\label{eq:TopologyPres}
\begin{aligned}
E_{\text{tp}}(\textbf h)&=\int_\Omega \delta_{\textbf h}(\textbf{w})d\textbf w, \text { with}\\
\delta_{\textbf h}(\textbf{w})&:=
\left\{
\begin{array}{c l}
     e^{-| J_{\textbf h}(\textbf w)|}+\varphi\sqrt{\vert J_{\textbf h}(\textbf w)\vert ^2}    &\text{if  }|  \; |J_{\textbf h}(\textbf w)| -1\;| \geq \tau \\
     0   & \text{otherwise}   \\
\end{array}\right.
\end{aligned}
\end{equation}
Here $\varphi \in  \mathbb{R}^+$ offers a balance in our penalization, and $\tau  \in  \mathbb{R}^+$ is the margin of acceptance for values close to one.

The first term
\[
	e^{-| J_{\textbf h}(\textbf w)|}
\]
heavily penalizes negative values of the deformation. Thus it prevents the field $\mathbf h$ from having distortions or penetration of boundaries. The term
\[
	\varphi\sqrt{\vert J_{\textbf h}(\textbf w)\vert ^2}
\]
with the parameter $\varphi$ controls the magnitude of the expansions and contractions.

Unlike most of the Jacobian determinant constrains, for example $\log(|J|)$ (see \cite{Ashburner::99}) and $e^{(|J|)}$ (as in \cite{Kybic::00}), we do not only guarantee the positivity of the Jacobian determinant but also enforce its value to stay near one. Moreover, we penalize big expansions in order to achieve more realistic deformations.

There are different options for solving the $L^{2}$-regularized class. Traditional methods include Gradient Descent, Newton's method, Nonlinear Conjugate Gradient, and Evolutionary Optimization Algorithms. However, they can get stuck at local minima, might need an infinite number of iterations to converge, or have a slow rate of convergence. A better alternative is the well-known Levenberg-Marquardt (LM) which offers better results with less computational time. For these reasons, in this work we use LM method
to minimize our energy functional.


\section{Experimental Results}
\label{sec:Experiments}
This section describes in detail the experiments that we conducted to validate our proposal.

\subsection{Subjects and acquisition}

\begin{figure}[!t]
\centering
\includegraphics[width=12.5cm]{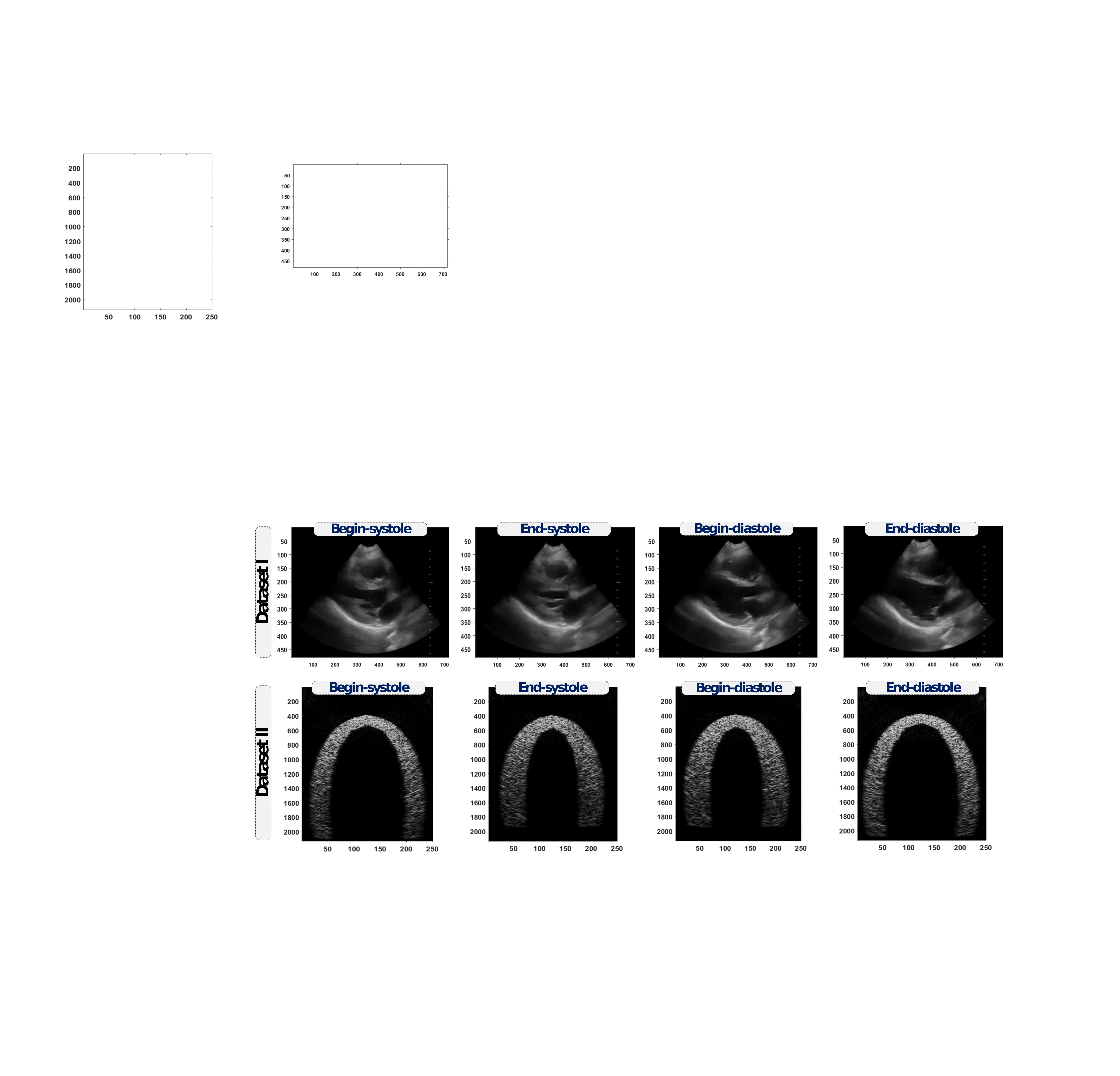}
\caption{Sample frames of the raw data extracted from the two datasets used for evaluating our approach.}
\label{fig:rawDataset}
\end{figure}

We used two ultrafast ultrasound datasets (see Figure~\ref{fig:rawDataset}) to evaluate our proposed approach.

The first is a realistic dataset from one patient. During the acquisition, the patient
was placed in the supine position or left lateral decubitus. Then, the probe was placed on the left parasternal line at the fourth intercostal space with the marker pointing toward the right shoulder of the patient. The images were thus taken from the parasternal long axis view. This view is useful for global assessment of the motion of the heart's wall and the function of different areas including the right and left ventricle, the mitral and aortic valves and the interventricular septum. This dataset is composed of  $1814$ frames with size of $720$x$480$, and a scaled version of size $144$x$96$.
This data was acquired with an UUS device with a (fc) transducer with a bandwidth of 6MHz and 192 elements using coherent compounding of plane waves. The output is a 2D plane wave image sequence of the long axis view of a healthy heart.

The second dataset (see~\cite{Nillesen::15} for details) is a simulated ultrafast ultrasound sequence that has a realistic cardiac deformation field and describes the mechanics of a healthy left ventricle (see \cite{Bovendeerd::09}  for the description of the mechanics).  This data was generated using (fc) transducer with a frame rate of 5000 Hz (single transmits) and effective frame rate of 500 Hz.
The output of this simulated data is a set of 2D apical imaging planes. This view is helpful to study the left ventricle and the mitral inflow. The sequence is composed of  $399$ frames with size of $2143$x$250$. This simulated data is very helpful to assess the performance of our approach since it includes a ground truth of the displacements which can be compared against our estimation.

All the measurements and reconstructions in this section are taken from these two ultrafast ultrasound cardiac sequences. All test and comparison were run under the same condition on a  CPU-based implementation. We used an Intel(R) Core i7- 6700 CPU at 3.40GHz-32GB RAM, and a Nvidia GeForce GT 610.

\begin{table*}[!t]
\centering
\label{table:preprocessing}
\resizebox{17cm}{!} {
\bgroup
\def\arraystretch{1.2}
\begin{tabular}{|l|l|l|l|l|l|l|l|l|}
\hline
\multicolumn{1}{|c|}{\textbf{Preprocessing}} & \multicolumn{1}{c|}{\textbf{\begin{tabular}[c]{@{}c@{}}Topology Regularizer\\ Our Approach\end{tabular}}} & \multicolumn{1}{c|}{\textbf{Minimum}} & \multirow{4}{*}{} & \multicolumn{1}{c|}{\textbf{\begin{tabular}[c]{@{}c@{}}Topology Regularizer\\ Rohlfing et al. \cite{Rohlfing::2003} \end{tabular}}} & \textbf{Minimum} & \multirow{4}{*}{} & \multicolumn{1}{c|}{\textbf{\begin{tabular}[c]{@{}c@{}}Topology Regularizer \\ Heyde et al. \cite{Heyde::2016}\end{tabular}}} & \textbf{Minimum} \\ \cline{1-3} \cline{5-6} \cline{8-9}
\textbf{Low-Rank (SVD)} & \multirow{3}{*}{\begin{tabular}[c]{@{}l@{}} $E_{\text{tp}}(\textbf h)=\int_\Omega \delta_{\textbf h}(\textbf{w})d\textbf w$ \\ with $\delta_{\textbf h}(\textbf{w})$ as in Eq. [14] \end{tabular}}  & $2.0357e^{-12}$ &  & \multirow{3}{*}{$E_{\text{tp}}(\textbf h)=\int_\Omega |log(J_{\textbf h}(\textbf{w}))| d\textbf w$} & $4.0028e^{-6}$ &  & \multirow{3}{*}{$E_{\text{tp}}(\textbf h)=\int_\Omega |J_{\textbf h}(\textbf{w})-1|^2 d\textbf w$} & $3.2012e^{-7}$ \\ \cline{1-1} \cline{3-3} \cline{6-6} \cline{9-9}
\textbf{Wavelet} &   & $6.0596e^{-6}$ &  &  & $9.6984e^{-4}$ &  &  & $1.6443e^{-5}$ \\ \cline{1-1} \cline{3-3} \cline{6-6} \cline{9-9}
\textbf{Gaussian Smoothing} &  & $1.7417e^{-3}$ &  &  &  $7.1822e^{-2}$ &  &  & $3.3613e^{-2}$  \\ \hline
\end{tabular}\egroup}
\caption{Performance comparison between our proposed approach  and other state of the art approaches}
\end{table*}

\subsection{Validation scheme}

We divided our validation scheme into two parts. The first part makes use of the realistic dataset and relies on the following measurements to evaluate the performance of our topology-preserving technique:
\begin{itemize}
\item Comparison between our proposed topology regularizer and two from the literature: Table 3;
\item Assessment of using low-rank as a preprocessing step: Figure \ref{fig::convergence};
\item Inspection of the displacement field:  Figure~\ref{fig:Distortions} (A)/(B);
\item Numerical results offered by the Jacobian determinant:  Figure~\ref{fig:Distortions} (A.2)/(B.2); Table 4;
\item Careful comparison of the residual error for both the low-rank tool and the topology preservation tool and study of their synergy: Table 4;
\end{itemize}


In the second part, we use a simulated dataset with a provided ground truth in which we perform the following evaluations:

\begin{itemize}
\item  Numerical visualization and comparison of the displacement field: Figures 8 and 9;
\item  Numerical visualization of mean accumulated displacement of the seven segments of the left ventricle  Figure~\ref{fig:accumulatedMeanSegments};
\item  Nonparametric statistical analysis between the real and the estimated displacements;
\item  Illustration of the computed strain as a reasonable clinical measure:   Figure~\ref{fig:strainPlots}.
\end{itemize}

\medskip

\subsection{Results}

In order to prove the benefits of using low-rank (SVD) as a preprocessing step, we compared it against two common preprocessing techniques: Gaussian smoothing (kernel $5x5$ and $\sigma=0.7$) and Wavelets (Biorthogonal Spline Wavelet, 4 levels). We carried out the comparison of the three preprocessing techniques using our topology regularizer and two more from the state of the art \cite{Rohlfing::2003,Heyde::2016}.

\begin{figure*}[!t]
  \centering
  \includegraphics[width=13cm]{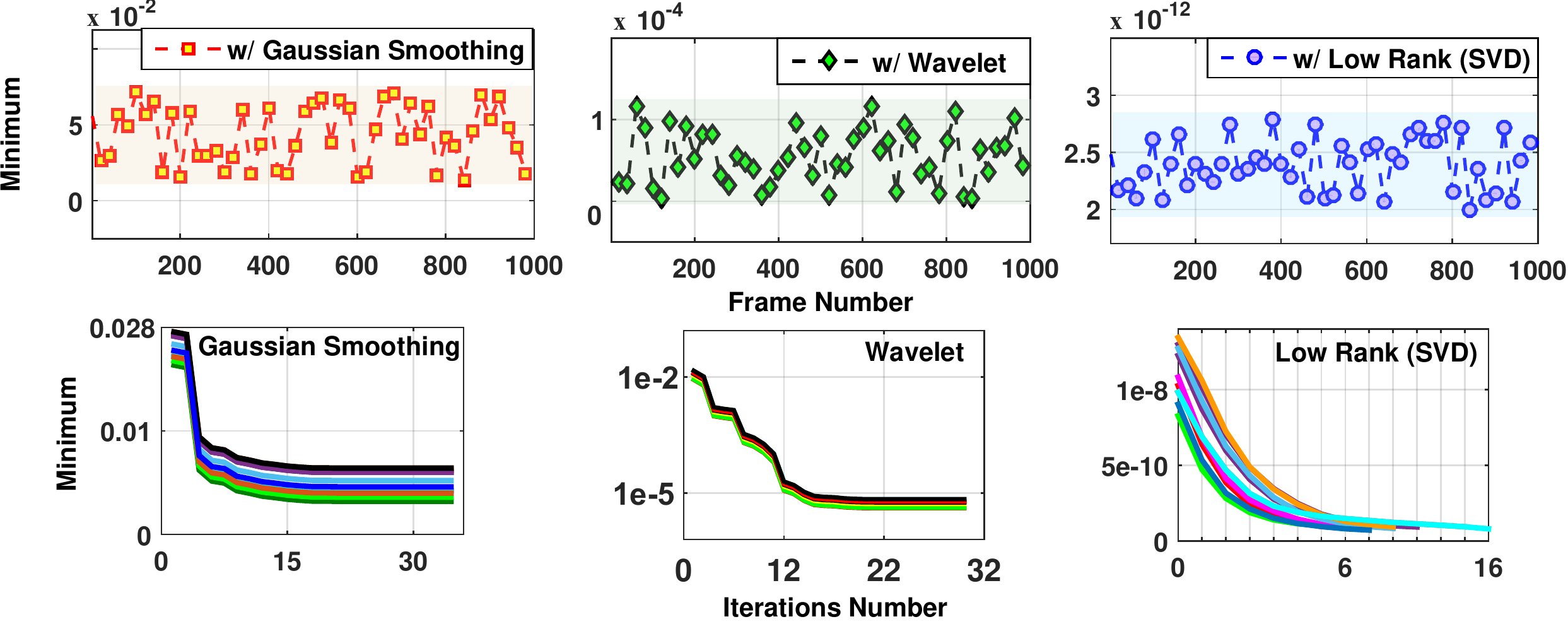}
  \includegraphics[width=4cm]{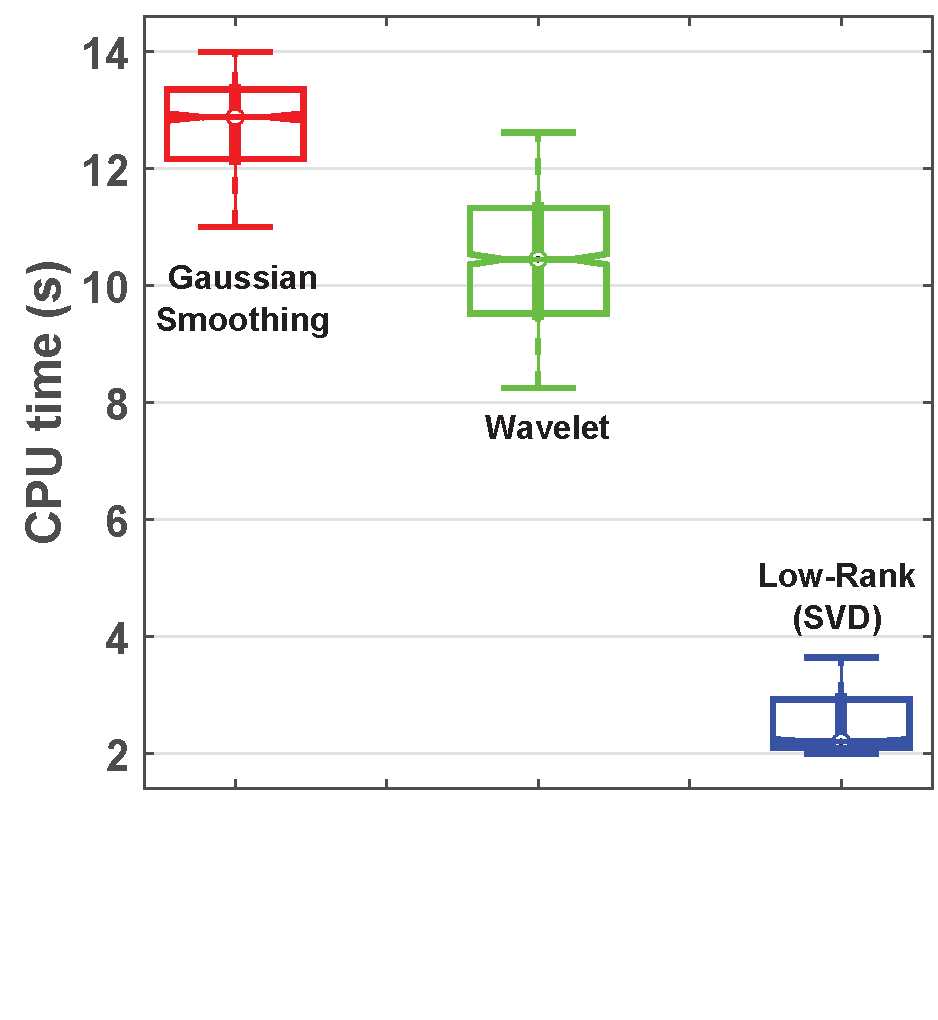}
  \caption{The first row shows the convergence history of the complete sequence (1000 frames) using three different pre-processing techniques, while the second row shows the number of iterations it took to find the minimum for few of those frames. Box-plot at the right side shows the CPU time comparison of the different techniques.}
  \label{fig::convergence}
\end{figure*}

According to the results (Table 3 and Figure \ref{fig::convergence}), we found that low-rank was able to find the best minima in our case study in a computationally efficient manner. The results showed that Wavelet was able to find an acceptable minima but, it needed an average of 30 iterations per frame to converge compared to the 16 needed by low-rank (see plots in Figure \ref{fig::convergence}). Gaussian smoothing on the other hand performed the worst in terms of minima and average iterations per frame.

Overall, out of the three prepossessing techniques, low-rank offered a good tradeoff between accuracy and computational time since it requires less iterations per frame. This is further reflected in the overall CPU time as illustrated in the box-plot at right side of Figure \ref{fig::convergence} where low-rank only needed an average of 2.3217 seconds of computational time while Gaussian and Wavelet both required more than 11 seconds in average. Moreover, it performed the best across the different regularizers giving the best minima each time. In general, our topology regularizer approach was the best compared to the regularizers offered by Rohlfing~\cite{Rohlfing::2003} and Heyde~\cite{Heyde::2016}. It has the advantage of enforcing the value to be close to one and penalizing very strong expansions and contractions.

\begin{table*}[!t]
\centering
\label{tab:Performance}
\resizebox{17cm}{!} {
\begin{tabular}{|cl||c|c|c|c|c|}
\hline
 \textbf{Exp.}& \textbf{Data} &
 \multicolumn{1}{|c|}{\begin{tabular}[c]{@{}c@{}}\textbf{Energy functional in \eqref{eq:Energy}:}\\ \textbf{specifics of topology preservation}\end{tabular}}
    &   \multicolumn{1}{|c|}{\begin{tabular}[c]{@{}c@{}}\textbf{CPU time (seconds)}\\ \textbf{per frame}\end{tabular}}  &  \textbf{Minimum}  & \multicolumn{1}{|c|}{\begin{tabular}[c]{@{}c@{}}\textbf{Discrepancy}\\ \textbf{(mean)} \eqref{eq:SSD}\end{tabular}} & \multicolumn{1}{c|}{\begin{tabular}[c]{@{}c@{}}\textbf{Average {[}min max{]}} \\\textbf{of} $|J_{h}(w)|$ \end{tabular}}
  \\ \hline
\multicolumn{7}{|c|}{}   \\
\hline
1)&  Full-Rank 	&	$E_{tp} = 0$	&  12.0830 & 0.1069  &   \multirow{4}{*}{0.0016}    &  [-2.8896 3.6884]
 \\  \cline{1-5} \cline{7-7}
 2)& Full-Rank 	&	 \eqref{eq:TopologyPres} with $\varphi=0$ 	&  12.0672  &  $3.9945e^{-3}$ &     & [0.0617 1.0096]
\\ \cline{1-5} \cline{7-7}
3)&Full-Rank 	&	\eqref{eq:TopologyPres} with $\varphi=10^{-2}$	&  11.3953 &  $6.0155e^{-3}$ &     & [0.9034 1.0023]
 \\ \cline{1-5} \cline{7-7}
4)&Full-Rank 	&	 \eqref{eq:TopologyPres} with $\varphi=5\cdot 10^{-3}$	& 11.0304   & $4.5563e^{-4} $ &    & [0.9546 1.0059]
\\
\hline
\multicolumn{7}{|c|}{}   \\
\hline
 5)&Low-Rank 	&	$E_{tp} = 0$	&  3.8771  &  $0.1191$ &   \multirow{4}{*}{7.9505e{-06}}      &[-1.0144 3.7375]
 \\ \cline{1-5} \cline{7-7}
 6)&Low-Rank 	&	\eqref{eq:TopologyPres} with $\varphi=0$	& 3.4895  &  $2.1994e^{-09} $ &      &[0.0213 1.0031]
\\ \cline{1-5} \cline{7-7}
7)&Low-Rank 	&	\eqref{eq:TopologyPres} with $\varphi=10^{-2}$	&  3.0214   & $3.2975e^{-11} $  &      &[0.9184 1.0028]
\\ \cline{1-5} \cline{7-7}
8)&Low-Rank 	&	\eqref{eq:TopologyPres} with $\varphi=5\cdot 10^{-3}$	& 2.32170   & $2.0357e^{-12} $  &       &[0.9950 1.0100]
 \\ \hline
\end{tabular}}
\caption{Performance analysis: low-rank  vs. full-rank data and their reaction to different degrees of topology preservation (see text for discussion).}
\end{table*}

\begin{figure*}[!t]
  \centering
\includegraphics[width=0.8\textwidth]{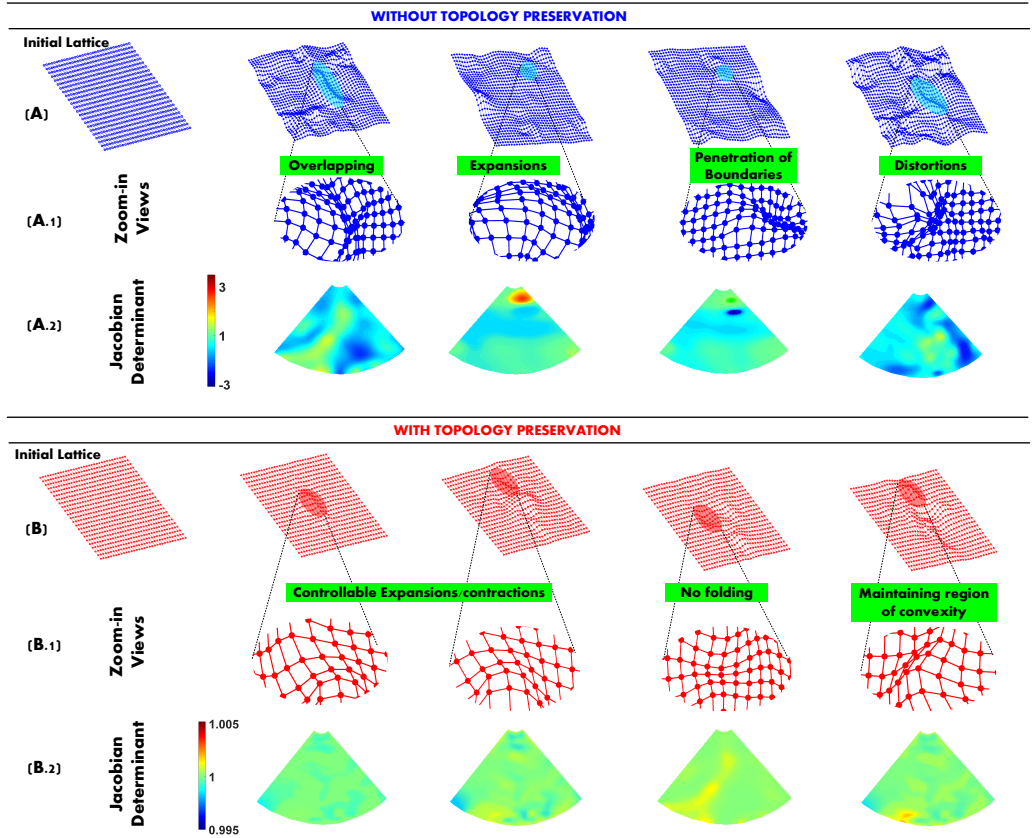}
\caption{(A) Resulted transformations, during complex deformations, without applying topology preservation. Highlighted areas denote structure violations that are more clearly displayed in the zoom-in views (A.1). The resulted Jacobian determinant are shown in (A.2). Resulted transformations, after applying topology preservation, are shown in (B) and can be compared with (A) in which (B.1) and (B.2) show that they keep the mesh structures with most of the Jacobian determinant staying at 1.}
\label{fig:Distortions}
\end{figure*}

The performance of our proposed topology-preserving technique is illustrated in Figure \ref{fig:Distortions} with and without topology preservation. First we show the performance without topology preservation  ($E_{tp}=0$ in the energy functional \eqref{eq:Energy}) where the resulting displacement fields are displayed in (A) and the corresponding Jacobian determinant are shown in (A.2). The example frames in the columns show critical details of the respective displacements fields and Jacobian. Row (A.1) shows zoom-in parts where different violations of the topology occur in part (A), such as overlapping, boundaries penetration, and mesh elements distortion.

The plots of the Jacobian clearly show huge variations in the value of the determinant which results in an unstable representation of the anatomical structure. Not only that, but in some parts the Jacobian determinant presents big values (greater than 3), which indicates that some transformations produced very big expansions, while in others the Jacobian determinant gave negative values, which indicates that new structures were formed.
These violations create new structures and result in an unrealistic representation of the complex deformation of the heart's motion. This is not acceptable particularly in medical applications where preservation of the anatomical structure is remarkably needed.

\begin{figure*}[!t]
\centering
\includegraphics[width=0.88\textwidth]{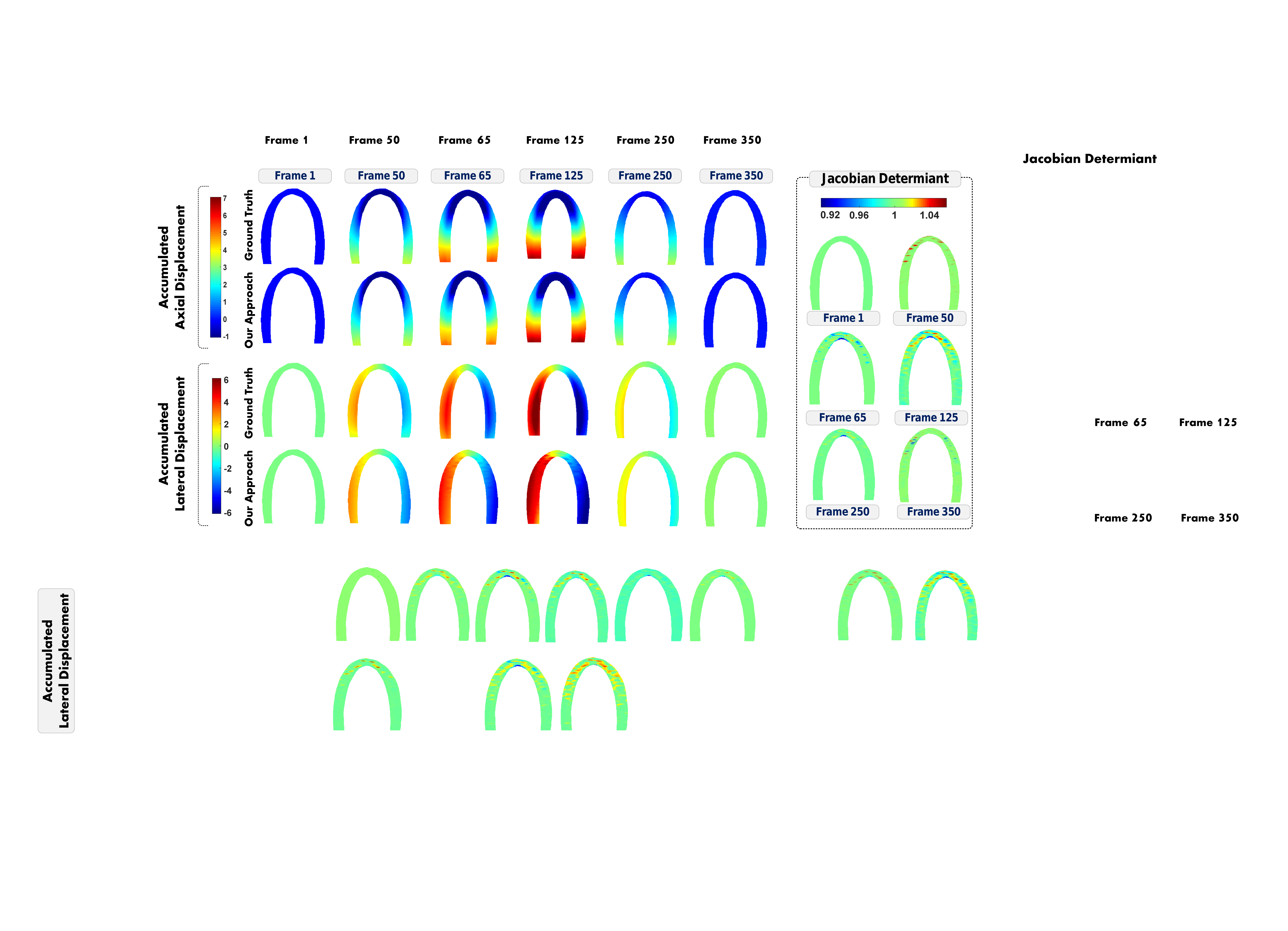}
\caption{ (From right to left) Accumulated displacement for the apical view of the left ventricle. Few samples of the approximated axial and lateral displacements (top and bottom) are compared against the ground truth.  Left side shows the Jacobian determinant of the same sample frames which reflects preservation of the anatomy.}
\label{fig:accumulatedDisplacement}
\end{figure*}

We then ran the same tests after applying the proposed topology preserving approach ($E_{tp}$ in \eqref{eq:TopologyPres} with $\varphi=5\cdot 10^{-3}$) and show the results in part B. Looking at the zoom in parts in B.1, and comparing it with part A.1, we can verify that our approach allows controlling expansions and contractions, maintaining region convexity, and avoiding foldings. The corresponding plots of the Jacobian determinant are shown in part B.2. In comparison to the Jacobians without topology preservation, we can see stabilities on the values as they do not suffer large variations (mostly stay on 1) with guaranteed positivity. These are realistic values, as over two pairs of time-frames, the volume should be approximately preserved. From these illustrations one can therefore conclude that the approach was successful in avoiding topology violation even with complex deformations of the cardiac motion.



For a more detailed quantitative analysis, we evaluated the global performance of our approach by comparing its performance with low-rank and full-rank data using $600$ frames of the sequence (see Figure~\ref{fig:rawDataset} dataset I). Unlike works where speckles were useful as indicators for tracking the heart's motion, in the framework we proposed, results from Table 4 show that promoting low-rank offered a positive effect on the solution as it significantly reduced the computational time and allowed faster convergence of the energy functional (less iterations per frame).


\begin{figure*}[!t]
\centering
\includegraphics[width=0.9\textwidth]{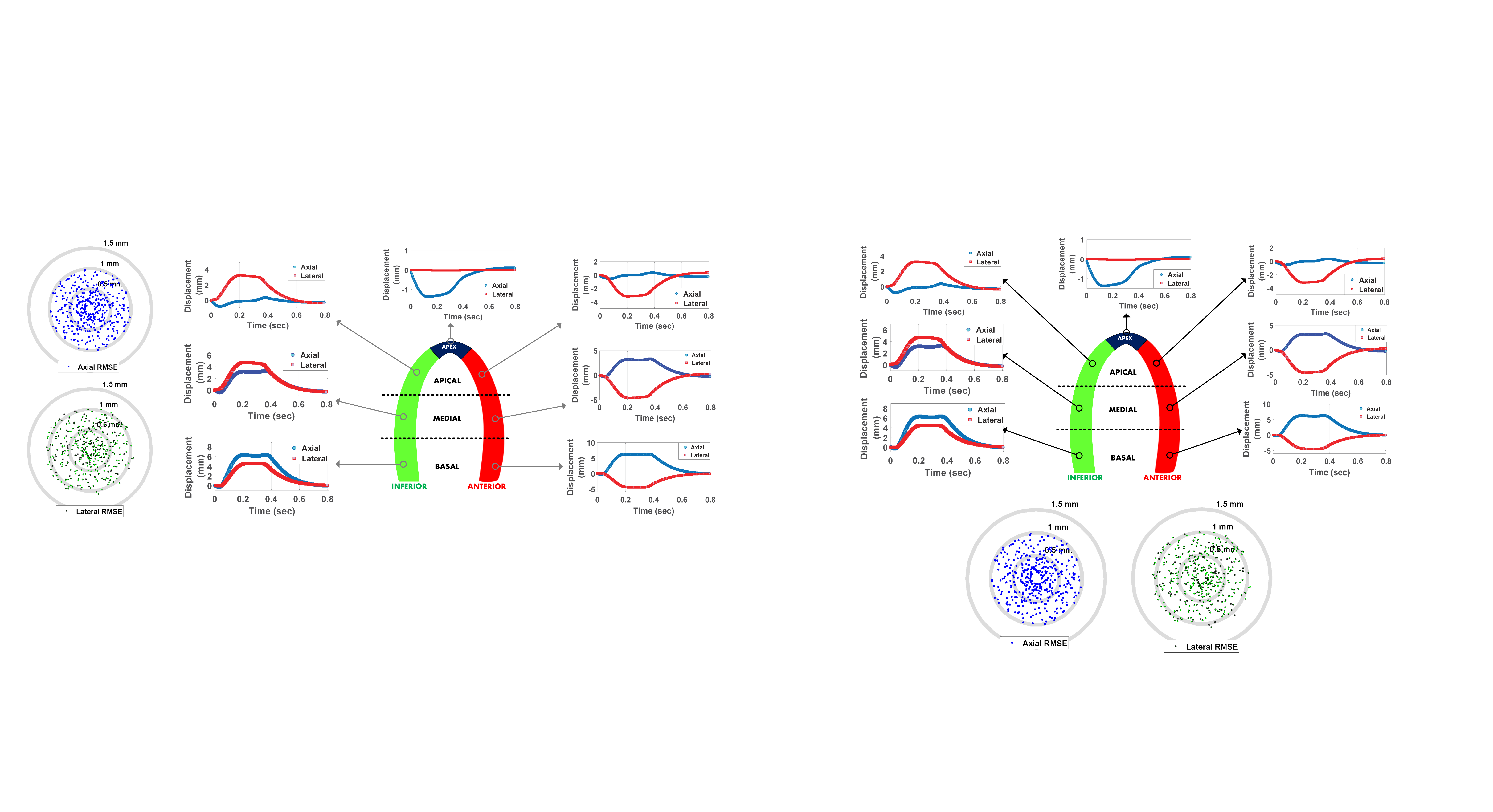}
\caption{Mean accumulated displacement of the seven segments of the left ventricle. Blue circles make reference to the axial displacement while red squared refer to the lateral displacement.}
\label{fig:accumulatedMeanSegments}
\end{figure*}

\begin{figure}[!t]
\centering
\includegraphics[width=0.59\textwidth]{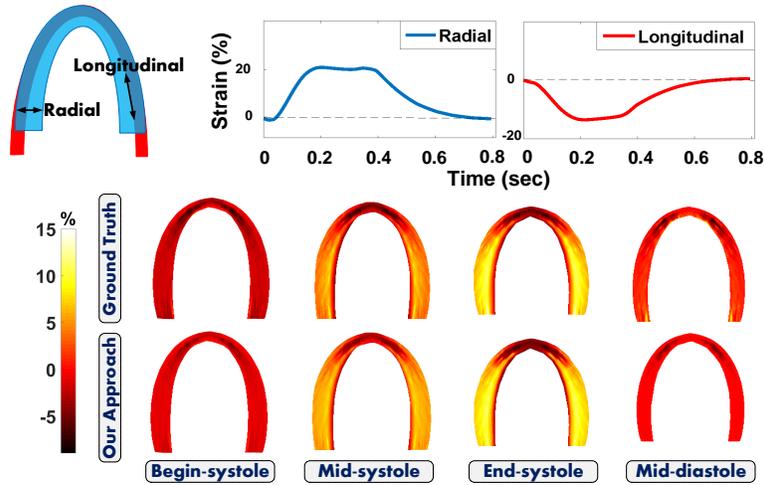}
\caption{(Top) Radial and longitudinal strain profiles of the left ventricle. These profiles are evaluated by their sign where negative values reflect shortening and positive ones reflect stretching. (Bottom) Few frames of the cardiac cycle showing the radial strain.  }
\label{fig:strainPlots}
\end{figure}

Table 4, Exp. 1 and Exp. 5 show that without topology preservation in the functional \eqref{eq:Energy}, the residuum was about $0.1$ with and without the low-rank constraint. Applying low-rank, the Jacobian determinant had higher values but still suffered from heavy distortions. With topology preservation in Exps. 2-4, and 6-8,  reasonable values of the Jacobian determinant were obtained, avoiding any penetration of boundaries. Thus, topology preservation is a necessary tool to get  realistic deformation results for cardiac motion estimation.

The primary role of the low-rank representation seems to be the radical decrease in computational time, as seen from a comparison of Exps. 1-4 and Exps. 5-8, where the computational time was decreased by about 75~\%.

The residual errors in Exps. 2-4 in the topology-preserved full-rank case were in the order of magnitude $10^{-3}$ resp. $10^{-4}$. Contrary to that, topology preservation in the low-rank case in Exps. 6-8 yielded minima in the order of magnitude $10^{-9}$ to $10^{-12}$.  Moreover, the discrepancy error showed that low-rank achieved an order of magnitude $10^{-6}$ in comparison with $10^{-3}$ given by the full-rank case. In practice, topology preservation and low-rank constraint act together to get a more realistic deformation field in less time.

As a second part of our validation scheme and for an extended evaluation of our proposal, we illustrate the axial and lateral accumulated displacement of both the ground truth and our estimation of a single heart cycle (left side of Fig. \ref{fig:accumulatedDisplacement}).  It is clear by visual inspection of the colored bar that our estimation is very close to the ground truth. In order to support this statement, we computed the Root-Mean Square error (RMSE) for the axial and lateral displacement and plotted the results in left side of Figure 9  where we can see RMSE values less than 1mm for the axial direction and less than 1.2mm for the lateral direction. The plots also show a concentration of values much lower than 1mm  in both displacement directions.

To complement the analysis of the estimated displacement, in right side of Figure \ref{fig:accumulatedMeanSegments} we offer an analysis of the seven segments of the left ventricle during one cycle of the heart. Since the inferior and anterior parts are symmetric, we can evaluate their behavior for the axial and lateral displacements.  As expected, axial displacements (blue circle) reported positive values acting in a similar way in both sides while lateral displacements (red squared) showed similar behavior with contrary signs since they go to opposite directions during  heart motion.

We used the nonparametric Wilcoxon signed rank sum test to answer whether there is statistical significant difference between the real values (ground truth) and the estimated values. We found that the null hyphotesis was not rejected with $p<0.05$ significance level. This, lead us to conclude that we obtained a good estimation of the displacement.


To further support the results obtained in Figure \ref{fig:Distortions} – (A.1 and A.2) and Table 4, the right side of Fig. \ref{fig:accumulatedDisplacement} shows the corresponding Jacobian determinant of the illustrated frames where we can see that most of the values meet the desired criteria (stay at 1) for achieving topology preservation, which proofs the efficiency of our proposed term.

Finally, we provide \emph{strain images} as these are often used clinically. It is well-known that the strain can be calculated in terms of the components of the displacement vector field$\text{ }\textbf h(\textbf w)$:
\[
	\varepsilon(\textbf w)=\begin{pmatrix}\varepsilon_{\textbf{xx}}(\textbf w) &\varepsilon_{\textbf{xy}}(\textbf w) \\ \varepsilon_{\textbf{yx}}(\textbf w) & \varepsilon_{\textbf{yy}}(\textbf w) \end{pmatrix}=
	\frac{1}{2} (C-\mathbf{I})=\frac{1}{2}\left(\mathbf{F}^{T}\mathbf{F}-\mathbf{I})\right),
\]
where $C$ is the Green deformation tensor, $\mathbf{F}=\nabla \mathbf h(\mathbf w)$ is the displacement gradient, and $\mathbf{I}$ is the identity.


Strain is useful to evaluate the heart muscle and to identify subtle changes in heart function \cite{Abraham::01}. Moreover, it allows representing the percentage change in dimension from a resting state to a stressed state (after applying a force). Figure \ref{fig:strainPlots} shows the radial and longitudinal strains related to the left ventricle. The resulted plots can be evaluated according to the sign of the strain in which negative values indicate shortening and positive values denote stretching. According to the results at the upper part of the figure, radial strain reported a stretching behavior while longitudinal strain a shortening behavior. For illustration purposes, the radial strain of few frames from the sequence are displayed at the lower part of the same figure.
The strain profiles and displacements of the left ventricle exhibit distinct features and clinically meaningful motion patterns.

\section{Conclusion}
\label{sec:Discussion}

In this paper, we presented a new approach to estimate the cardiac motion using ultrafast ultrasound data. In a variational framework, we combined a penalizer for topology preservation with a low-rank data representation. Together with the better temporal resolution of ultrafast ultrasound, our proposed approach overcame challenges of non-rigid registration, including noise and complex heart motion and inaccurate results exhibiting distortions. While keeping the computational time relatively low, a realistic and clinically meaningful displacement field was produced, with the diffeomorphic features and preserved structures.

In our variational framework, the displacement was represented by a lattice with splines, and a maximum likelihood estimator was used in the discrepancy term to provide robustness against outliers. The regularizer for the topology has two features:  eliminating radically negative values and carefully controlling the volume expansion and compression. We validated the accuracy of our approach and showed that it offers a RMSE lower than 1 mm in comparison to the ground truth.

While this variational framework gives good results and is strong individually, the consuming CPU time and artifacts in the ultrafast ultrasound sequence motivated us to promote a low-rank data representation, as it has proved to be useful in other areas of imaging and computer vision. We represented the data in a single Casorati matrix and used the dominant singular values to compute the deformation. Apart from removing the noise in the ultrasound data, this technique  greatly reduces the computational time and produces, together with the topology penalization term, significantly less discrepancy in the results.

While we wanted to show the potentials of combining ultrafast ultrasound with low-rank techniques and topology preservation from a technical point of view, the objective of this work is to have a first study for proof of concept and to open a new line of research for further clinical investigation. In this work, we evaluated the technique using data from one subject. Future work will include a more extensive evaluation with more subjects to examine the clinical potentials of the approach. Moreover, the technique promises to be useful for analyzing organs experiencing complex motion other than the heart, as for example the movement of the lungs in respiration.

\section*{Acknowledgment}
We would like to thank Chris de Korte for discussion and coordination and Peter Bovendeerd for the initial code for the simulated data.
Support from the ERC Advanced Grant Project MULTIMOD 267184 is greatly acknowledged.  Also, support from a FPU national scholarship with reference AP2012-1943. \\

\end{document}